\documentclass[conference]{IEEEtran}
\IEEEoverridecommandlockouts
\usepackage{cite}
\usepackage{amsmath,amssymb,amsfonts}
\usepackage{algorithmic}
\usepackage{graphicx}
\usepackage{textcomp}
\usepackage{xcolor}
\usepackage{multirow}
\usepackage{hyperref}

\def\BibTeX{{\rm B\kern-.05em{\sc i\kern-.025em b}\kern-.08em
    T\kern-.1667em\lower.7ex\hbox{E}\kern-.125emX}}
\begin{document}

\title{Sentimental LIAR: Extended Corpus and Deep Learning Models for Fake Claim Classification\\
}

\author{\IEEEauthorblockN{Bibek Upadhayay}
\IEEEauthorblockA{\textit{SAIL Lab} \\
\textit{University of New Haven}\\
West Haven, CT, USA \\
bupad1@unh.newhaven.edu}
\and
\IEEEauthorblockN{Vahid Behzadan}
\IEEEauthorblockA{\textit{SAIL Lab} \\
\textit{University of New Haven}\\
West Haven, CT, USA \\
vbehzadan@newhaven.edu}
}
 
\maketitle

\begin{abstract}
The rampant integration of social media in our every day lives and culture has given rise to fast and easier access to the flow of information than ever in human history. However, the inherently unsupervised nature of social media platforms has also made it easier to spread false information and fake news. Furthermore, the high volume and velocity of information flow in such platforms make manual supervision and control of information propagation infeasible. This paper aims to address this issue by proposing a novel deep learning approach for automated detection of false short-text claims on social media. We first introduce Sentimental LIAR, which extends the LIAR dataset of short claims by adding features based on sentiment and emotion analysis of claims. Furthermore, we propose a novel deep learning architecture based on the BERT-Base language model for classification of claims as genuine or fake. Our results demonstrate that the proposed architecture trained on Sentimental LIAR can achieve an accuracy of 70\%, which is an improvement of ~30\% over previously reported results for the LIAR benchmark.
\end{abstract}


\section{INTRODUCTION}
The percolation of social media throughout the world has facilitated unprecedented ease of access to the flow of information. The rise of the internet and its availability have also enabled every user to to not only consume, but also contribute to the information flow. However, the benefits of such ecosystems come at the cost of mistrust in the veracity of information. In recent years, the social media scene has witnessed the proliferation of false information campaigns, in which ordinary users are intentionally or otherwise both consuming false news and also spreading it among their communities. 

This phenomenon is commonly referred to as \emph{fake news}, broadly defined as broadcasting of information that is intentionally and verifiably false \cite{zhou-zafarani}. The rise of fake news and its societal impact has been studied in the context of numerous recent events, such as the Brexit referendum and the 2016 US presidential elections \cite{Pogue.2017}. Fake news has thus proven to be a major threat to democracy, journalism, and freedom of expression \cite{zhou-zafarani}. The exposure of users to fake news has been shown to have numerous deleterious effects, instances of which include inducing attitudes of inefficacy, alienation, trusting in false propaganda, cynicism toward certain political candidates and communities, that can at times give rise to the violent events. For example, coordinated fake news and propaganda campaigns on Facebook are considered to have been key in inciting the Myanmar genocide in 2016-2017 \cite{Myanmar.2016}. Also, the recent proliferation of false information about 5G communication networks being the cause of the novel Coronavirus outbreak has resulted in attacks against the employees and infrastructure of cellular careers in the UK \cite{laato2020people}. Fake news can also affect financial markets, as observed in the case of fake news claiming that Barack Obama was injured in an explosion resulting in a loss of \$130 billion in stock value \cite{Rapoza.2017}. Hence, there is a growing need for effective tools and techniques to detect and control the spread of false information campaigns on social media. 

Fake news classification is the process of determining whether the news contains false news and misinformation or not. Traditionally, this classification is performed by subject-matter experts and journalists via comparing the claims of an article with established facts and cross-checking with trusted and alternative sources. However, the high volume and velocity of information flow on such platforms render such manual approaches infeasible. Therefore, recent efforts of the stakeholders and the research community have been focused on automated techniques for classification and detection of fake news. A promising solution in this domain is to leverage the recent advances in machine learning and Natural Language Processing (NLP) to automated the processing and classification of the high-dimensional and complex text of news articles and posts \cite{nlp-survey}. 
%

While the literature on the applications of machine learning to fake news classification has grown rapidly, the body of work on the classification of short-text claims remains relatively thin. This issue is of paramount importance, as many of the posts on social media such as Twitter contain only a short claim extracted from the longer text of news articles. The short form of such claims poses a challenge to the classification task, as it provides very limited information (i.e., a few sentences or words) and thus constrains the applicability of machine learning models trained on full-length articles and texts. Over the past few years, a number of datasets and models have been proposed for the classification of short-text claims, notable instances of which are the studies based on the LIAR dataset of short statements \cite{wang2017liar}. However, the performance of machine learning models trained on this dataset remain at impractical levels, with the best accuracy values reported to be \~41.5\% \cite{nlp-survey}. 



In this paper, we introduce Sentimental LIAR, which extends the LIAR dataset by including new features based on the sentiment and emotion analysis of claims. Our extended dataset also proposes a modified encoding of textual attributes to mitigate unintended bias in modeling. Furthermore, we propose a novel deep learning architecture based on the BERT-Base language model for the classification of claims as genuine or fake. Our results demonstrate that the proposed architecture trained on Sentimental LIAR can achieve an accuracy of 70\%, which is an improvement of ~30\% over previously reported results for the LIAR benchmark. The Sentimental LIAR dataset and the proof-of-concept code are made available on GitHub\footnote{\url{https://github.com/UNHSAILLab/SentimentalLIAR}}.


The remainder of this paper is organized as follows: Section \ref{sec:literature} presents the technical background and an overview of relevant datasets and literature on false claim classification. Section \ref{sec:methodology} describes the extended features of Sentimental LIAR, and details the proposed deep learning architectures for false claim detection. The experimental evaluation of our proposed techniques is reported in Section \ref{sec:experiment}. Finally, \ref{sec:conclusion} concludes the paper with a discussion on the results and remarks on future directions of work.

\section{LITERATURE OVERVIEW}
\label{sec:literature}
\subsection{Fake News}
\emph{Fake news} is defined as ``fabricated information that mimics news media content in form but not in organizational process or intent'' \cite{Lazer.2018}. Fake news outlets exploit the fact that social media platforms lack the editorial norms and processes of the traditional news media for assuring the accuracy and credibility of the information. Fake news overlaps with other information disorders, such as misinformation (false or misleading information) and disinformation (false information that is purposely spread to deceive people)" \cite{Lazer.2018}. Fake news can also be defined as news that is false based on its authenticity (false or not), intention (bad or not), and whether the information is news or not \cite{zhou-zafarani}. Undeutsch hypothesis \cite{undeutsch1967beurteilung} implies that fake news are different from the true news in terms of their writing style. And the four-factor theory \cite{zuckerman1981verbal} implies that the fake statements are expressed with different emotions and sentiments than the truth ones. These two theories support the intuition that the identification of the attributes like emotions and sentiments in a claim can help to distinguish between the fake and true claims

Technical approaches to the detection of fake news include fact checking, rumor detection, stance detection, and sentiment analysis \cite{nlp-survey}. \emph{Fact checking} is the task of assessing the truthfulness of claims made by public figures such as politicians and pundits \cite{vlachos2014fact}. The contents of fake news often lacks pertinent facts, or contain factual representations that are not correct according to the context of the news. \emph{Rumor} can be defined as the unverified pieces of information at the time of posting, where there is doubt to the truth of the claims. Zubiaga et al. \cite{Zubiaga_2016} define rumor detection as the task of separating personal statements into rumor or non-rumor. \emph{Stance detection} refers to the process of automatically detecting whether the author of a piece of text is in favor of the given claim or against it \cite{krejzl2017stance}. \emph{Sentiment analysis} is based on extracting emotions and contextual sentiments from statements to determine whether they are positive, negative or neutral with respect to the subject. It is noteworthy that while stance detection may rely on sentiment analysis, the goal of sentiment analysis is to analyze personal emotions rather than the objective verification of statements. As detailed in Section \ref{sec:methodology}, our work leverages a hybrid of stance detection and sentiment analysis to enhance the performance of machine learning models for false claim detection in short text.


\subsection{DATASETS}
Recent research on fake claim detection in short text has yielded a number of significant open-source datasets, some of the most notable of which are enumerated as follows:

\subsubsection{FEVER \cite{Thorne_2018}} Fact Extraction and VERification (FEVER) is a short claim dataset of 185,445 claims. This dataset is annotated with three labels: \emph{Supported, Refuted,} and \emph{Not Enough Info}. The claims are curated from Wikipedia. FEVER was constructed in two stages, the first one is Claim Generation, where extracted information Wikipedia is converted into claims. The second stage is Claim Labeling, where each claim is labeled as 'supported' or 'refuted' according to Wikipedia. In cases where information were insufficient for determination, a label of \emph{not enough information} is assigned.

\subsubsection{PHEME \cite{Zubiaga_2016}} is a dataset of 330 rumor threads composed of 4843 tweets associated with 9 newsworthy events. The annotators of this dataset were journalists who tracked the events in real time. Each entry in PHEME is labeled as either true or false. 

\subsubsection{LIAR \cite{wang2017liar}}  
\begin{figure}[h!]
\centering
  \includegraphics[width=0.5\textwidth]{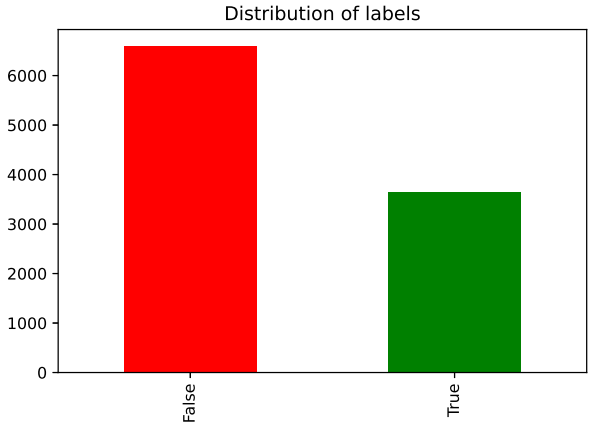}
  \caption{Label Distribution in LIAR}
  \label{fig:LabelDist}
\end{figure}

is a publicly available short statement dataset that is derived from Politifact.com. Each of the 12,836 statements in LIAR is annotated based on data available on Politifact with one of the following six labels: pants-fire, false, barely true, half-true, mostly-true, and true. The dataset contains the text of a claim, as well as relevant meta-data, structured as follows: ID, LABEL, STATEMENT, SUBJECT, SPEAKER, SPEAKER JOB, STATE INFO, PARTY AFFILIATION, BARELY TRUE COUNTS, FALSE COUNTS, HALF TRUE COUNTS, MOSTLY TRUE COUNTS, PANTS ON FIRE COUNTS, and CONTEXT. The distribution of the labels in the LIAR dataset is illustrated in Fig.\ref{fig:LabelDist}.


\subsection{Deep Learning for Fake Claim Detection}
 Convolutional Neural Networks (CNNs) are generally known for their applications in image processing, where CNNs can learn representations of localized features from raw input while preserving the relationship between them. However, this capability of CNNs is not constrained to computer vision, and can also be adopted for NLP tasks. Wang et al. \cite{wang2017liar} demonstrate the performance a CNN architecture \cite{KIMCN} trained on LIAR for fake claim classification, and report a test-time accuracy of 27.4\%. In a subsequent study, Kirilin et al. \cite{kirilin2018exploiting} extend the LIAR dataset with a larger credibility source using a credibility vectorization technique called Speaker2Credit. Experimental results reported in \cite{nlp-survey} demonstrate that training a Long Short-Term Memory (LSTM) model on this extended dataset achieved an accuracy of 45.7\%.

Also, Long et al. \cite{long-etal-2017-fake} propose another extension of the LIAR dataset by adding more features based on speaker profiles, such as party affiliation, speaker title, location, and credit history. They also report that an attention-based LSTM model trained on this extended dataset results in an improved accuracy of 41.5\% compared to the benchmark of 27.4\% reported in \cite{wang2017liar}.

Ruchansky et al. \cite{Ruchansky_2017} propose a hybrid deep model for fake news detection, consisting of three modules: The first module is a Recurrent Neural Network (RNN) that captures the temporal pattern of user activity on a given article, the second module learns the article's source characteristic based on the user response, and the third module integrates the previous modules for the classification task. Their study was based on two datasets curated from Twitter and Weibo, and achieved an accuracy score of 89.2\% and 95.3\%, respectively. 


 \subsection{BERT}
 Bidirectional Encoder Representations from Transformer (BERT) \cite{BERT-DBLP:journals/corr/abs-1810-04805} is a state of the art word representation model that uses a transformer to learn the contextual relationships between the given text in a bidrectional manner (i.e., in both left-to-right and right-to-left). The bidirectionality of BERT has made it standout in many NLP tasks, as it improves fine-tuning based approaches to token level tasks. BERT uses two strategies in training, the first one is Masked Language Model (MLM) and second one is Next Sentence Prediction (NSP). In MLM, the model randomly masks some of the tokens from the input text and then model tries to predict those masked vocabulary id of the word. It does so by looking into the context in both direction. As for Next Sentence Prediction, the pairs of sentences are given input to the model and it tries to predict if the given second sentence is the continuation or the next subsequent sentence to the first one or not. The model uses both MLM and NSP during training to minimize the loss. BERT can be modified to perform many downstream tasks by feeding the task-specific inputs and outputs. For the classification tasks, BERT can be extended by adding a classification layers on the top of the transformer output for the token.

\section{METHODOLOGY}
\label{sec:methodology}
Fake claims are often written in a style of exaggerated expressions and strong emotions. Style-based classification studies aim to assess news intention, that is to determine whether there is an intention to mislead the reader or not? The fake claims are written with an intention to convince the audiences to read and trust the claims, for which fake claims are written with different styles \cite{undeutsch1967beurteilung} and different sentiments and emotions \cite{zuckerman1981verbal}.  With the aim of developing a computationally feasible model for fake claim detection, we propose deep neural network architectures based on BERT-Base to analyze the deception in short-text claims.  Our proposed models learn to detect deception based on attribute-based language features, such as sentiments and structure-based language features. In our approach, we rely on the representation learning capabilities of transformers to extract features from statements. We also propose an extension to the LIAR dataset that includes additional features based on the sentiment and emotion analysis of the claim. The details of our proposals are presented in the remainder of this section.

\subsection{Sentimental LIAR}

Our Sentimental LIAR dataset is a modified and further extended version of the LIAR extension introduced by Kirilin et al. \cite{kirilin2018exploiting}. In our dataset, the multi-class labeling of LIAR is converted to a binary annotation by changing half-true, false, barely-true and pants-fire labels to False, and the remaining labels to True. Furthermore, we convert the speaker names to numerical IDs in order to avoid bias with regards to the textual representation of names. 

The binary-label dataset is then extended by adding sentiments derived using the Google NLP API\footnote{\url{https://cloud.google.com/natural-language}}. Sentiment analysis determines the overall attitude of the text (i.e., whether it is positive or negative), and is quantified by a numerical score. If the sentiment score is positive, then we assign \emph{Positive} for the sentiment attribute, otherwise \emph{Negative} is assigned. We also introduced a further extension by adding emotion scores extracted using the IBM NLP API\footnote{\url{https://www.ibm.com/cloud/watson-natural-language-understanding}} for each claim, which determine the detected level of 6 emotional states, namely anger, sadness, disgust, fear and joy. The score for each emotion is between the range of 0 and 1. Table \ref{tab:LIARexample} demonstrates a sample record in Sentimental LIAR for a short claim in the LIAR dataset. 




\begin{table}[]
\begin{tabular}{|c|l|l|}
\hline
\multirow{7}{*}{TEXT} & statement & \begin{tabular}[c]{@{}l@{}}McCain opposed a \\ requirement  that the \\ government buy American\\ -made motorcycles. And\\ he said all buy-American\\ provisions were quote\\  'disgraceful.'\end{tabular} \\ \cline{2-3} 
 & subject & federal-budget \\ \cline{2-3} 
 & speaker\_id & \_2\_ \\ \cline{2-3} 
 & speaker\_job & President \\ \cline{2-3} 
 & state\_info & Illinois \\ \cline{2-3} 
 & party\_affiliation & democrat \\ \cline{2-3} 
 & sentiment & NEGATIVE \\ \hline
\multirow{5}{*}{EMO} & anger & 0.1353 \\ \cline{2-3} 
 & disgust & 0.8253 \\ \cline{2-3} 
 & sad & 0.1419 \\ \cline{2-3} 
 & fear & 0.0157 \\ \cline{2-3} 
 & joy & 0.0236 \\ \hline
\multirow{5}{*}{SPC} & barely\_true\_counts & 70 \\ \cline{2-3} 
 & false\_counts & 71 \\ \cline{2-3} 
 & half\_true\_counts & 160 \\ \cline{2-3} 
 & mostly\_true\_counts & 163 \\ \cline{2-3} 
 & pants\_on\_fire\_counts & 9 \\ \hline
\multicolumn{1}{|l|}{SEN} & sentiment\_score & -0.7 \\ \hline
\end{tabular}
\caption{\label{tab:LIARexample}Sample Record from Sentimental LIAR}

\end{table}

\subsection{Models}
We investigate two model architectures based on BERT-Base \cite{BERT-DBLP:journals/corr/abs-1810-04805} for the claim classification task: 

\subsubsection{Model 1: BERT-Base with Feed-Forward Neural Network}
\begin{figure}[h!]
\centering
  \includegraphics[width=0.5\textwidth]{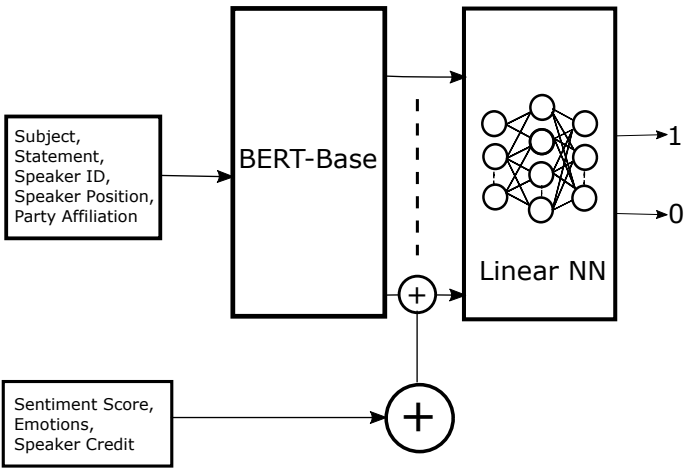}
  \caption{\label{fig:DB_NN}BERT-Base with feed-forward component for classification}
\end{figure}

This model extends BERT-Base by appending a Feed-Forward neural network for classification, as illustrated in fig.(\ref{fig:DB_NN}). Two variations of the model were investigated: The first variant feed all the input data directly to the BERT-Base, the second variant passes only the TEXT input to BERT-Base, and feeds the SEN, EMO and SPC attributes in parallel to the output of BERT-Base to the feed-forward component, as depicted in fig.(\ref{fig:DB_NN}).

\subsubsection{Model 2: BERT-Base with CNN}
\begin{figure}[h!]
\centering
  \includegraphics[width=0.5\textwidth]{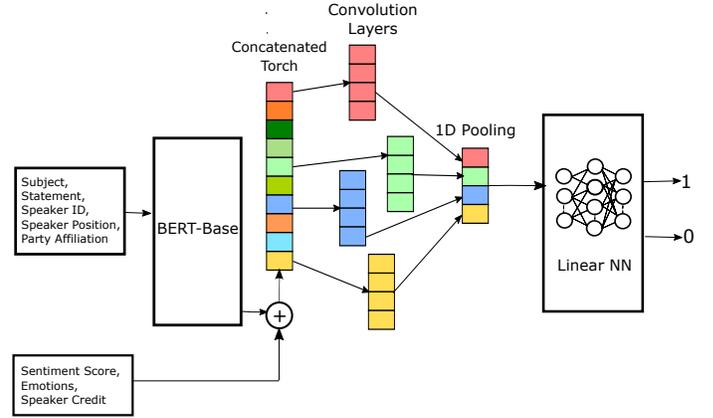}
  \caption{\label{fig:DB_CNN}BERT-Base with CNN for Classification}
\end{figure}

The base model is modified by appending a CNN to BERT-Base, as shown in fig.(\ref{fig:DB_CNN}). Similar to the previous model, we investigate two variants of this architecture. In first variant, the TEXT, SPC, SEN and EMO are fed into the BERT-Base, whose output is then passed to the CNN. In the second variant, only the TEXT is fed into BERT-Base, and the EMO, SPC and SEN attributes are concatenated with output of BERT-Base to be fed into the CNN.

\section{Experimental Results}
\label{sec:experiment}
In both  experiments, the dataset was split into 80\% train set, 10\% valid set, and 10\% test set. The train batch size and the test batch size were set to 8, with a learning rate of 1e-05. The BERT-Base was configured with a dropout of 0.3 and model used Sigmoid as its activation function. The loss function used in both architectures was the binary cross entropy loss optimized using the Adam optimizer \cite{kingma2014adam}.

\subsection{Experiments with BERT-Base + Feed-Forward Neural Network}

The first method was used in three experiments with BERT-Base + Feed-Forward NN. The model was composed of BERT-Base layers, a dropout layer, and one feed-forward hidden layer. The TEXT and EMO attributes were first fed into model and the output of BERT-Base (BB\_OP) was fed into the feedforward component. This model achieved the accuracy score of 64.92\% with a F1 Score of 0.6105. The input to BERT-Base was then extended by adding more meta data: TEXT+EMO+SPC and TEXT+EMO+SPC+SEN, yielding an accuracy of 67\% and F1 Score of 0.40.

The model-1 was then modified based on the second variant. The hidden layers in neural nets were increased to size six (with dimensions 768, 800, 512, 256, 128, 2). The TEXT attribute was fed into BERT-Base and the BERT-Base's output (BB\_OP) was concatenated with the EMO, SPC, SEN attributes to be passed to the feedforward NN. The model performed better than the previous variant, with the accuracy score of 69.37\% and the F1 Score of 0.57234. The accuracy and F1-score for the experiments performed with BERT-Base with feed-forward NN are given in Table \ref{tab:DB_NN}.

\begin{figure}[h!]
\centering
  \includegraphics[width=0.5\textwidth]{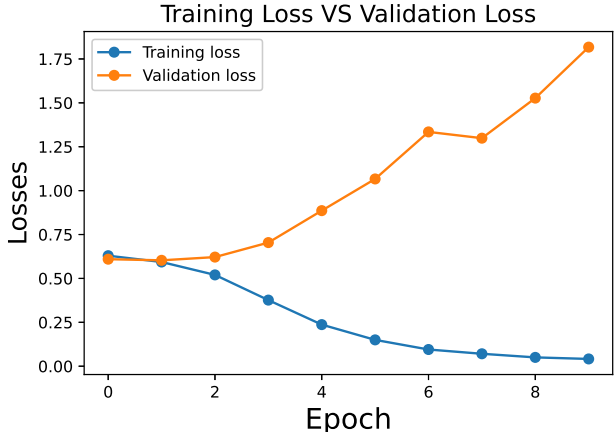}
  \caption{\label{fig:graph-NN}Experiment 5 Training Loss VS Validation Loss}
\end{figure}

\begin{table}[]
\begin{tabular}{|l|l|l|l|}
\hline
S.N. & Experiment & Accuracy & \begin{tabular}[c]{@{}l@{}}F1 Score\\ Macro\end{tabular} \\ \hline
1. & \begin{tabular}[c]{@{}l@{}}TEXT $ \rightarrow $ {}{[}BB{]},\\ BB\_OP $ \rightarrow $ {}{[}NN{]}\end{tabular} & 0.6882 & 0.5842 \\ \hline
2. & \begin{tabular}[c]{@{}l@{}}TEXT+EMO  $ \rightarrow $ {}{[}BB{]},\\ BB\_OP $ \rightarrow $  {[}NN{]}\end{tabular} & 0.6773 & 0.6352 \\ \hline
3. & \begin{tabular}[c]{@{}l@{}}TEXT+EMO+\\ SPC $ \rightarrow $ {}{[}BB{]},\\ BB\_OP $ \rightarrow $ {}{[}NN{]}\end{tabular} & 0.6720 & 0.4021 \\ \hline
4. & \begin{tabular}[c]{@{}l@{}}TEXT+EMO+\\ SPC+SEN $ \rightarrow $ {}{[}BB{]},\\ BB\_OP $ \rightarrow $ {}{[}NN{]}\end{tabular} & 0.6734 & 0.4097 \\ \hline
5. & \begin{tabular}[c]{@{}l@{}}TEXT $ \rightarrow $ {}{[}BB{]},\\ BB\_OP+EMP+\\ SPC+SEN $ \rightarrow $ {}{[}NN{]}\end{tabular} & 0.6937 & 0.57234 \\ \hline
\end{tabular}
\caption{\label{tab:DB_NN}BERT-Base with Feed-Forward NN, Accuracy and F1 Score}
\end{table}

\subsection{Experiments with BERT-Base + CNN}

In second model, a CNN component was appended to BERT-Base with 2 1D convolution layers. The first layer input channel size was 1 and output channel size was 50, and the second layer input channel size was 50 and output channel size was 100. The kernel size for both layers was 20 with stride of 1. The 1D max-pooling of size one was used in all the experiments. We started with the first variant, where the TEXT and other attributes were fed directly into BERT-Base, the output (BB\_OP) of which was passed into the CNN. In our first experiment, only TEXT was given to BERT-Base, yielding an accuracy score of 68.82\% with a F1 Score of 0.5308. Both the accuracy and the F1 Score were similar to the best performer of model-I. In the next experiment, only TEXT was fed into BERT-Base and the output of BERT-Base was concatenated with EMO and then fed into CNN. The model produced an accuracy of 65.54\% and a F1 Score of 0.608. In the fourth experiment, the TEXT and SPC attributes were fed into BERT-Base and the output was concatenated with EMO to be passed into the CNN. This model performed better with the accuracy of 68.90\% and F1 Score of 0.6542.
In the fifth experiment, TEXT was fed into BERT-Base, and its output was concatenated with EMO and SPC before feeding into CNN. The model performed better than all others, yielding an accuracy of 70\% and F1 Score of 0.630. The accuracy and F1-score for the experiments performed with BERT-Base + CNN are given in Table \ref{tab:DB_CNN}.

\begin{figure}[h!]
\centering
  \includegraphics[width=0.5\textwidth]{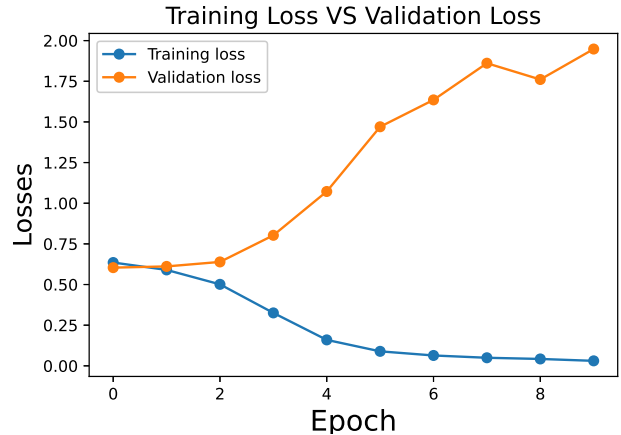}
  \caption{\label{fig:graph-CNN}Experiment 6, Training Loss VS Validation Loss}
\end{figure}

\begin{table}[]
\begin{tabular}{|l|l|l|l|}
\hline
S.N. & Experiment & Accuracy & \begin{tabular}[c]{@{}l@{}}F1 Score\\ Macro\end{tabular} \\ \hline
1. & \begin{tabular}[c]{@{}l@{}}TEXT $ \rightarrow $ {}{[}BB{]},\\ BB\_OP $ \rightarrow $ {}{[}CNN{]}\end{tabular} & 0.6882 & 0.5308 \\ \hline
2. & \begin{tabular}[c]{@{}l@{}}TEXT+EMO +\\ SPC $ \rightarrow $ {}{[}BB{]},\\ BB\_OP $ \rightarrow $  {[}CNN{]}\end{tabular} & 0.5546 & 0.55641 \\ \hline
3. & \begin{tabular}[c]{@{}l@{}}TEXT $ \rightarrow $ {}{[}BB{]},\\ BB\_OP+\\ EMO $ \rightarrow $ {}{[}CNN{]}\end{tabular} & 0.6554 & 0.608 \\ \hline
4. & \begin{tabular}[c]{@{}l@{}}TEXT+SPC $ \rightarrow $ {}{[}BB{]},\\ BB\_OP+\\ EMO $ \rightarrow $ {}{[}CNN{]}\end{tabular} & 0.6890 & 0.6542 \\ \hline
5. & \begin{tabular}[c]{@{}l@{}}TEXT $ \rightarrow $ {}{[}BB{]},\\ BB\_OP+EMO\\ +SPC $ \rightarrow $ {}{[}CNN{]}\end{tabular} & 0.7000 & 0.6370 \\ \hline
6. & \begin{tabular}[c]{@{}l@{}}TXT $ \rightarrow $ {}{[}BB{]},\\ BB\_OP+EMO+\\ SPC+SEN $ \rightarrow $ {}{[}CNN{]}\end{tabular} & 0.6992 & 0.6430 \\ \hline
\end{tabular}
\caption{\label{tab:DB_CNN}BERT-Base + CNN, Accuracy and F1 Score}

\end{table}

\section{Discussion and Conclusion}
\label{sec:conclusion}
This paper introduced Sentimental LIAR as an extension of the LIAR dataset, and proposed novel model architectures based on BERT-Base for fake claim detection in short text. The proposed architectures extend BERT-Base by adding (1) a feedforward neural network, or (2) a CNN. The LIAR dataset is extended by adding emotions anger, sad, fear, anger and disgust by using IBM NLP API and added sentiment score using Google NLP API. We also included speaker credit as an input attribute to our models. 

The experiments performed with BERT-Base + feedforward NN, the accuracy ranged from 68.8\% to 69\% within the five experiments. These experiments were performed by changing the input structure in the first three experiments and by changing the hidden layers in the latter two experiments. A slight improvement of 1\% was observed in the accuracy and no improvements in the F1 Score. This suggests that the model may need to be revised to handle the complexity of the input data. Hence, a CNN-based architecture was investigated in our further experiments. 

The experiments were performed with BERT-Base + CNN, the accuracy ranged from 68.82\% to 70\%  within six experiments, and also major improvements were observed in the F1 Score (0.5308 to 0.6430). The best performing model is found to be one where the text attribute is fed directly into BERT-Base, and the output of BERT-Base is concatenated with the emotions, speaker's credit and sentiments before being passed to the CNN. Undeutsch hypothesis \cite{undeutsch1967beurteilung} and the four-factor theory \cite{zuckerman1981verbal} supported the intuition that the emotional and sentimental attributes can help to distinguish the fake claims, which can be verified by the observation of the model performing better when EMO and SEN were added. Adding the SEN and EMO with BERT-Base output supplemented the features which boosted the CNN model performance. 

For both models, it can be observed that adding the metadata (i.e., emotions, sentiments, and speakers' credit) increased the accuracy of model. Also, both the model accuracy and the F1 Score improved with the CNN-based architecture. 

The training loss VS validation Loss graphs for BERT-Base + feedforward NN is given in Fig.(\ref{fig:graph-NN}), and for BERT-Base + CNN in Fig.(\ref{fig:graph-CNN}). These plots suggest that the models were overfitted only after 2 epochs, which is mostly due to the small size of the dataset. Also, it must be noted that the dataset is imbalanced, with 65\% of data labeled as false and only 35\% labeled as true. These observations demonstrate the need for the curation of larger and more representative datasets of short-text claims.

Furthermore, our results further verify that fake claims can be detected in short-text according to exaggerated expressions and strong emotions demonstrated in the text. The proposed architecture also sets a new state-of-the-art benchmark for fake claim classification on the LIAR dataset with an accuracy of 70\%.


\bibliographystyle{IEEEtran}
\bibliography{ref}
\vspace{12pt}

\end{document}